\documentclass{article}

\usepackage{arxiv}

\usepackage[utf8]{inputenc}
\usepackage[T1]{fontenc}
\usepackage{hyperref}
\usepackage{url}
\usepackage{booktabs}
\usepackage{amsfonts}
\usepackage{nicefrac}
\usepackage{microtype}

\usepackage{graphicx}
\usepackage{natbib}
\usepackage{doi}
\usepackage{lipsum}
\usepackage{amssymb}
\usepackage{amsmath}
\usepackage[normalem]{ulem}
\usepackage{multirow}
\usepackage{array}
\usepackage{bm}
\usepackage{caption,subcaption}
\usepackage{pifont}
\usepackage{xcolor}
\usepackage{makecell}
\usepackage{longtable}

\usepackage{fancyhdr}
\fancypagestyle{plain}{
\fancyfoot{}
\fancyfoot[L]{\footnotesize \copyright 2026. This accepted version is made available under the CC-BY-NC-ND 4.0 license. The definitive version of record has been published in \textit{Expert Systems with Applications (Elsevier), 2026}, and is available at: \url{https://doi.org/10.1016/j.eswa.2026.131173}}
\fancyfoot[R]{\thepage}

}

\title{Fusing Memory and Attention: A study on LSTM, Transformer and Hybrid Architectures for Symbolic Music Generation
\thanks{\footnotesize \copyright 2026. This accepted version is made available under the CC-BY-NC-ND 4.0 license. The definitive version of record has been published in \textit{Expert Systems with Applications (Elsevier), 2026}, and is available at: \url{https://doi.org/10.1016/j.eswa.2026.131173}}}

\author{
    Soudeep Ghoshal \\
    Kalinga Institute of Industrial Technology (KIIT)\\
    Bhubaneswar, India \\
    \texttt{2205421@kiit.ac.in} \\
    \And
    Sandipan Chakraborty \\
    Kalinga Institute of Industrial Technology (KIIT)\\
    Bhubaneswar, India \\
    \texttt{2205412@kiit.ac.in} \\
    \And
    Pradipto Chowdhury \\
    Kalinga Institute of Industrial Technology (KIIT)\\
    Bhubaneswar, India \\
    \texttt{21051913@kiit.ac.in} \\
    \And
    Himanshu Buckchash \\
    IMC University of Applied Sciences Krems\\
    Krems, Austria \\
    \texttt{himanshu.buckchash@imc.ac.at} \\
}

\date{}

\renewcommand{\headeright}{}
\renewcommand{\undertitle}{}
\renewcommand{\shorttitle}{Fusing Memory and Attention}

\hypersetup{
    pdftitle={Fusing Memory and Attention: A study on LSTM, Transformer and Hybrid Architectures for Symbolic Music Generation},
    pdfauthor={Soudeep Ghoshal, Sandipan Chakraborty, Pradipto Chowdhury, Himanshu Buckchash},
    pdfkeywords={Deep Learning, LSTM, Transformer, Symbolic Music Generation, Sequence Modeling},
    colorlinks=true,
    linkcolor=black,
    citecolor=black,
    urlcolor=blue
}

\begin{document}
\maketitle

\begin{abstract}
Machine learning techniques, such as Transformers and Long Short-Term Memory (LSTM) networks, play a crucial role in Symbolic Music Generation (SMG). Existing literature indicates a difference between LSTMs and Transformers regarding their ability to model local melodic continuity versus maintaining global structural coherence. However, their specific properties within the context of SMG have not been systematically studied. This paper addresses this gap by providing a fine-grained comparative analysis of LSTMs versus Transformers for SMG, examining local and global properties in detail using 17 musical quality metrics on the Deutschl dataset. We find that LSTM networks excel at capturing local patterns but fail to preserve long-range dependencies, while Transformers model global structure effectively but tend to produce irregular phrasing. Based on this analysis and leveraging their respective strengths, we propose a Hybrid architecture combining a Transformer Encoder with an LSTM Decoder and evaluate it against both baselines. We evaluated 1,000 generated melodies from each of the three architectures on the Deutschl dataset. The results show that the hybrid method achieves better local and global continuity and coherence compared to the baselines. Our work highlights the key characteristics of these models and demonstrates how their properties can be leveraged to design superior models. We also supported the experiments with ablation studies and human perceptual evaluations, which statistically support the findings and provide robust validation for this work.
\end{abstract}

\keywords{Deep Learning \and LSTM \and Musical Structure Analysis \and Sequence Modeling \and Symbolic Music Generation \and Transformer}

\section{Introduction} \label{section:introduction}

Melody synthesis can be referred to as the generation of musical sequences using computational approaches. Given the \textbf{sequential nature} of music, the study requires models capable of capturing both local level note patterns and overarching musical structures. The challenge of effective melody synthesis lies in choosing architectures that can model intricate \textbf{temporal dependencies} while maintaining coherence over long musical phrases. While deep learning models have significantly advanced this field, the selection of an appropriate sequence modeling architecture remains an open question. \textbf{Recurrent Neural Networks (RNNs)}, especially \textbf{Long Short-Term Memory (LSTM)} networks, have traditionally dominated this space due to their strength in modeling sequential patterns \citep{shahid2022music,yu2020conditional}. However, their sequential processing nature can impede their ability to capture \textbf{long-range dependencies}, often resulting in degraded performance on complex compositions \citep{ou2023investigating}. On the other hand, \textbf{Transformer architectures}, with their \textit{self-attention mechanisms}, have demonstrated remarkable success in modeling global dependencies in natural language and are increasingly being explored in music generation \citep{shih2022theme, muhamed2021symbolic}. Yet, their ability to capture low-level rhythmic nuances is not always consistent.

Despite these advances, a critical research gap remains: the systematic comparison of how memory-based architectures (LSTMs) versus attention-based architectures (Transformers) fundamentally differ in musical generation. Existing literature provides fragmented evidence of complementary strengths. \citet{zheng2023comparative} demonstrated that LSTMs excel at local coherence in short sequences while Transformers capture long-term dependencies in extended passages through attention mechanisms. This raises a fundamental question: Can LSTM's local variance capture be reconciled with Transformer's global structural modeling to address the inherent tension between modeling local melodic continuity and maintaining global structural coherence? This tension manifests measurably through local variance metrics (pitch variance, interval variability, rhythmic patterns) reflecting immediate melodic behaviors \citep{huron2008sweet}, versus global variance metrics (entropy measures, motif diversity, harmonic complexity) assessing structural properties across entire compositions \citep{pearce2006expectation}. Yet no prior work has systematically investigated whether hybrid architectures combining LSTM's sequential memory with Transformer's attention mechanisms can simultaneously optimize both local and global musical characteristics.

Several past studies have attempted to apply these architectures in \textbf{symbolic music modeling}. For instance, Transformer-based approaches like Music Transformer \citep{huang2018musictransformer} and LSTM-based models \citep{simon2017performance} have been used independently to generate polyphonic and monophonic melodies. While these efforts have contributed valuable insights, a comprehensive comparative study analyzing their core architectural behaviors, especially through the lens of hybridization, remains largely unexplored. A right deep learning framework addressing this synthesis is therefore considered as potential direction of further research since most of the prior work focuses on isolated models without making systematic comparisons.  For these reasons, this paper, critically evaluates the properties of LSTMs vs. Transformers for SMG, using 17 local and global measures.

It is important to note that this study \textbf{prioritizes generative diversity and musical expressiveness over predictive accuracy}. In the field of symbolic music generation, \textit{high accuracy in prediction is usually associated with outputs that are more deterministic and that imitate the training patterns very closely} which, in turn, might restrain the productive exploration. Our assessment framework puts the emphasis on the balance between the soundness of the structure and the variability of the creativity, acknowledging that the right musical generation from models to be those which can create new but musically valid sequences rather than mere copies of the training data \citep{esling2022challenges}. To verify these ideas, we support the evaluations with appropriate selection of qualitative metrics.

In this work, we present a systematic study comparing three distinct architectures: a \textbf{pure LSTM model}, a \textbf{pure Transformer model}, and a proposed \textbf{hybrid model} that leverages the global pattern recognition capability of a Transformer encoder along with the temporal precision of an LSTM decoder. To further understand the influence of specific architectural components on melody generation, we conduct a series of \textbf{ablation experiments} on the hybrid model. These include modifications to the encoder depth, attention width, decoder structure, input representation, and regularization schemes. Our observations suggest that combining architectural strengths leads to more balanced melody generation, where local melodic continuity and global structure are better preserved. Certain modifications, such as simplifying the encoder or adjusting positional encoding, were found to subtly affect rhythmic stability and thematic consistency, revealing nuanced dependencies between architectural design and musical expressiveness. In addition to computational metrics, we also incorporate a \textbf{structured human perceptual evaluation} to bridge the gap between algorithmic assessment and subjective listening experience. This dual perspective ensures that the reported improvements reflect not only statistical robustness but also genuine enhancements in perceived musical quality.

The structure of this paper proceeds as follows: Section~\ref{section:related_works} examines existing literature on sequence modeling approaches for symbolic music generation. Section~\ref{section:methodology} outlines our methodology, detailing the model architectures and experimental framework. Section~\ref{section:results} analyzes the experimental results and explores significant findings. Finally, Section~\ref{section:conclusion} provides concluding remarks.

\section{Related Works} \label{section:related_works}

The Music Transformer, developed by \citet{huang2018musictransformer}, has taken a major stride in neural music generation through its novel use of relative positional self-attention mechanisms. While the architecture is based on the standard Transformer, this model overcomes key shortcomings in the language of long-term musical structures by implementing relative attention that encodes timing and pitch relations instead of absolute positions. The method enjoys massive improvements over various datasets, achieving state-of-the-art performance on JSB Chorales with a validation negative log-likelihood of just 0.335 against 0.417 of the baseline Transformer, and on Piano-e-Competition (1.835 vs. 1.969 of Performance RNN) against LSTM-based models. A more memory-efficient implementation of this model reduces the complexity to O(LD) from O(L²D) in computation, facilitating the generation of coherent minute-long compositions that exhibit remarkably strong internal consistency and generalize well beyond the lengths of training sequences. Yet, several obstacles and drawbacks confront Music Transformer. For very long sequences, the model remains quadratic in complexity despite memory improvements, which may indeed impose stringent constraints on extended musical forms. The Piano-e-Competition evaluation stymied only about 6 years of data ($\sim$1100 pieces), which may have stifled exposure to various piano styles and thus curtail generalizability.

\citet{singh2024explainabledeeplearninganalysis} devised a CNN-LSTM system for ARI in Hindustani Classical Music; this was a leap in computational musicology through employing deep learning architectures together with explainable AI techniques. To elaborate, convolutional neural networks were utilized in extracting local features, while LSTM layers processed faceted temporal sequences, which were pitch-based chromagram features with tonic normalization for aligning different performances to some common reference pitch. The framework is able to classify the 12 Raga classes with very high performance, with the CN2+LSTM+T configuration achieving the highest F1-score of 0.89, which considerably improved compared to the simpler architectures like CN1+T (0.63) which displays the utmost importance of tonic normalization since the one without such normalization, CN2+LSTM, could hardly even generalize (F1-score: 0.50). Two explanation methods validate the model; SoundLIME reaches 0.68 and GradCAM++ achieves 0.61 precision in tracing musically relevant regions within the time interval of one second. Despite this, there are some hurdles restricting the broad applicability of this framework. Firstly, the explainability analysis is limited to only 12 Raga classes, which considerably narrows down the scope for analysis; secondly, the model is not powerful enough to distinguish between closely related Ragas that overlap in musical features. Furthermore, the explainability mechanisms rely primarily on saliency-based techniques that may fail to capture nuanced musical ornamentations such as ``Khatka'' or ``Murki,'' potentially overlooking critical melodic subtleties that define Raga characteristics in Hindustani Classical Music traditions.

\citet{math11081915} established \textit{MelodyDiffusion}, the first transformer diffusion technique for chord-conditioned melody generation, utilizing transformers instead of U-Nets to capture long-distance dependencies in discrete musical sequences. The framework uses a BERT-like encoder to extract contextual chord information to condition melody generation for 500 diffusion timesteps with a schedule of Gaussian noise ($\beta$: 0.0001--0.02). While assessing its performance, architectural scaling showed remarkable improvements: the base one (12 layers, 768 hidden dimensions) reached Hits@1 accuracy of 70.35\%, whereas the large one (24 layers, 1024 hidden dimensions) attained 72.41\%, thus surpassing far an encoder-free approach (66.78\%) and competing with the Stable Diffusion baseline (71.78\%). However, critical drawbacks limit its practical application. Limiting the system to monophonic synthesis is an evident bottleneck in today's music generation, wherein polyphonic capabilities prevail. In addition, the quadratic attention complexity of the transformer imposes a high computational load for handling longer sequences, thus confining their application to brief segments of music. Training costs skyrocket with increasing sequence length, with the scene set for the inability of longer compositions owing to the lack of mechanisms for stringent temporal coherence. The framework's inability to handle polyphonic textures or concurrent melodic lines severely restricts its utility for realistic musical applications, demanding substantial architectural innovations to overcome these scalability and expressiveness challenges in future developments.

The multi-instrument music synthesis framework by ~\citet{hawthorne2022multiinstrumentmusicsynthesisspectrogram} sets forth an elaborate three-stage pipeline that pairs encoder-decoder Transformers with GAN-based spectrogram inversion and DDPMs for rendering a high-quality audio output from MIDI corresponding to various instrument configurations. The architecture converts MIDI inputs to spectrograms by means of stacked Transformer layers with self-attention; then, spectrogram inversion is realized via GANs employing 1D convolutions, transposed convolution layers, and residual blocks with dilated convolutions for audio synthesis. The ``Base w/ Context'' setup delivers a cakewalk to optimum performances on all metrics of evaluation: 3.13 and 1.00 reconstruction distances (VGGish and TRILL, respectively), Fréchet Audio Distance values of 2.74 and 0.27, respectively, a transcription F1-score of 0.36, and faster-than-real-time synthesis speeds. Having trained on large-scale datasets like MAESTROv3, Slakh2100, and MusicNet, the model is warranted to truly bring very flexible multi-instrument generation with greater audio quality than autoregressive counterparts. Some issues, though, stand in the way of such a model being truly useful: audio artifacts and loudness inconsistencies plague the shortcoming outputs, and there exists a rather gaping chasm of fidelity between generated audio and training audio. The spectrogram inversion stage represents a critical bottleneck, with GAN-based conversion introducing quality degradations that limit overall system performance. Secondly, smaller models without sufficient context encoding capacity find it hard to properly deal with segment transitions, which result in sounds discontinuities on longer compositions; finally, the computational complexity of the three-stage pipeline requires rather large resources with real-time scenario applicability in mind.

Recent advances in diffusion-based and variational autoencoder (VAE) architectures have introduced novel approaches to symbolic music generation, though they continue to grapple with the local-global coherence challenge. \citet{zhang2024composer} proposed a two-stage framework combining Vector Quantized VAE (VQ-VAE) with discrete diffusion models for composer-specific generation, achieving 72.36\% accuracy in style-conditioned generation. While this approach demonstrates effective global style modeling, the discrete representation limits fine-grained control over local melodic nuances. Similarly, \citet{mariani2023multi} introduced multi-source diffusion models capable of simultaneous music generation and source separation, representing the first unified model for both tasks. However, these diffusion based approaches, while excelling at capturing global distributional properties, face inherent limitations in modeling immediate local dependencies and maintaining micro-level melodic coherence, challenges that recurrent architectures traditionally address more effectively. This observation reinforces the fundamental research gap: existing advanced generative approaches continue to exhibit the same local-global trade-off that characterizes the LSTM-Transformer divide, suggesting that architectural hybridization may offer more balanced solutions than standalone diffusion or VAE frameworks.

\begin{table}[t]
\centering
\caption{Comparative Analysis of Existing Approaches: Local vs. Global Structure Modeling.}
\label{tab:lit_rev}
\resizebox{\textwidth}{!}{
\begin{tabular}{|p{1.8cm}|p{2.2cm}|p{3cm}|p{3cm}|p{3.5cm}|p{2.5cm}|}
\hline
\textbf{Model} & \textbf{Architecture} & \textbf{Local Structure Capability} &
\textbf{Global Structure Capability} & \textbf{Key Limitations} & \textbf{Dataset (s)} \\
\hline

Music Transformer \citep{huang2018musictransformer} &
Trans\-former with relative positional self-attention &
Maintains local timing grid through relative attention &
Strong: Generates minute-long compositions with coherent long-term structure; captures motifs and ABA form &
Quadratic complexity for long sequences; limited to \(\sim\)6 years of training data &
JSB Chorales, Piano-e-Competition \\
\hline

Perform\-ance RNN \citep{simon2017performance} &
LSTM-based &
Strong: Generates plausible short-term structure; maintains local continuity &
Weak: Loses long-term structure; forgets primer quickly; drifts to unrelated material &
Poor long-range dependency modeling; cannot extend user primers coherently &
Piano-e-Competition \\
\hline

CNN-LSTM for Raga \citep{singh2024explainabledeeplearninganalysis} &
CNN + LSTM with tonic normalization &
Strong: Extracts local features; F1-score 0.89 with tonic normalization &
Limited evaluation on global coherence &
Struggles with closely related Ragas; explainability restricted to 12 classes &
Hindustani Classical Music dataset \\
\hline

Melody Diffusion \citep{math11081915} &
Transformer-based diffusion model with BERT-like encoder &
Moderate: Hits@1 = 72.41\% &
Strong: Attention captures long-distance dependencies &
Monophonic only; quadratic attention limits long sequences; no polyphony &
Chord-conditioned melody generation dataset \\
\hline

Multi-Instrument Synthesis \citep{hawthorne2022multiinstrumentmusicsynthesisspectrogram} &
Encoder-decoder Transformers + GAN + DDPM &
Challenges in segment transitions for smaller models without context encoding &
Strong: Flexible multi-instrument generation with context modeling &
Audio artifacts, loudness issues; GAN-based inversion degrades quality; computationally expensive &
MAESTROv3, Slakh2100, MusicNet \\
\hline

VQ-VAE + Discrete Diffusion \citep{zhang2024composer} &
Vector Quantized VAE with discrete diffusion models &
Moderate: VQ-VAE compresses symbolic music into discrete codebook indexes &
Strong: Diffusion models trained on discrete latent space achieve 72.36\% accuracy in composer style generation &
Limited to symbolic music; lacks real-time controllability; computational overhead of two-stage generation process &
Classical composer datasets \\
\hline

Multi-Source Diffusion \citep{mariani2023multi} &
Diffusion-based generative model for joint probability density &
Strong: Handles source imputation (partial generation) with local context awareness &
Strong: First model capable of both music generation and separation; learns joint probability of sources sharing context &
High computational complexity; limited to source separation tasks; requires extensive training on multi-track data &
Slakh2100 \\
\hline

LSTM vs Transformer Study \citep{zheng2023comparative} &
Comparative analysis of LSTM and Transformer architectures &
LSTM Strong: Better local coherence and short-sequence structure &
Transformer Strong: Better long-term musical dependencies and extended passages &
LSTM: Loses coherence on longer works; Transformer: struggles with short-sequence structural recognition &
Not specified \\
\hline
\end{tabular}}
\end{table}

The comparative study performed by \citet{zheng2023comparative} addresses the basic structural problems inherent in neural music generation by systematically examining the capacity of LSTM and Transformer architectures to simulate coherent musical organization. Their approach involved a series of comparative experiments-testing on generated compositions for structural violations, ranging from short in length to long, and revealing crucial performance differences that help in the choice of architecture for music generation tasks. The testing showed an initial advantage for LSTMs in structural simulation for short sequences, thus favoring local coherence plus melodic continuity whereas a major drawback is shown in the loss of coherence in longer works because of gradients vanishing and restricted memory retention. Transformers instead possess the reverse capability; they first face difficulty in short sequence structural recognition, but they present a spectacular improvement when faced with extended music passages, as the enhanced attention mechanisms help them simulate long-term musical dependencies recognized by human beings. It is thus derived from the research that Transformer models are probably the best candidates on which to base AI music composition, once the Transformer attention mechanism has been improved to optimize structural pattern recognition. However, the study acknowledges limitations in current attention mechanisms' ability to capture hierarchical musical structures and suggests that future developments should focus on refining these mechanisms to better recognize compositional patterns across multiple temporal scales.

As evident from Table~\ref{tab:lit_rev}, the literature reveals a consistent pattern: memory-based architectures (LSTMs) demonstrate superior performance in maintaining local melodic continuity and short-term structural coherence, while attention-based architectures (Transformers) excel at capturing long-range dependencies and global structural patterns. However, no prior work has systematically investigated whether these complementary capabilities can be reconciled through hybrid architectures that strategically combine both mechanisms.

Building upon this work, our research extends their comparative framework by implementing architectures specifically designed to address the identified limitations in musical structure simulation. The following sections will elaborate on our  methodology, experimental design, and the resulting conclusions inferred from it.

\section{Methodology} \label{section:methodology}

\subsection{Dataset Selection and Justification}

This study utilizes the \textbf{``Deutschl''} dataset of the \textbf{Essen Folk Song Collection} \citep{schaffrath_essen} consisting of German folk melodies in \textbf{Kern format (.krn)}. The selection of this dataset was driven by methodological considerations designed to enable clean architectural comparison rather than stylistic objectives, addressing a critical challenge in music generation research: disentangling architectural performance from confounding stylistic variables.

The primary objective is to \textit{systematically compare how memory-based (LSTM) versus attention-based (Transformer) mechanisms} fundamentally differ in modeling local melodic continuity and global structural coherence. Introducing polyphonic textures or genre-diverse samples would conflate architectural effects with stylistic complexity, making it difficult to isolate whether performance differences stem from model architecture or genre-specific modeling challenges. For instance, jazz improvisation requires modeling complex harmonic progressions while classical polyphony demands simultaneous voice leading, which would introduce subjective evaluation criteria that obscure the core research question of memory versus attention mechanisms.

The Deutschl subset offers several methodological advantages: the Kern format provides \textbf{standardized symbolic representation} with consistent encoding essential for systematic comparison; the \textbf{manageable dataset size} enables comprehensive evaluation across 1,000 generated melodies per model while maintaining \textbf{computational feasibility for extensive ablation studies}; and the homogeneous structural characteristics (including consistent phrase lengths and relatively simple melodic contours) \textbf{minimize confounding variables}. This provides a controlled experimental environment analogous to using simplified datasets (just like MNIST in computer vision) to establish architectural principles before scaling to complex domains.

While this controlled approach limits immediate applicability to polyphonic or genre diverse tasks, it establishes fundamental architectural principles regarding the memory-attention trade-off that can inform future extensions. This methodological choice \textbf{prioritizes internal validity and interpretability of architectural effects over immediate external validity}, following established practices in controlled comparative studies. The observed patterns regarding local versus global modeling capabilities are expected to generalize as architectural properties, though their specific manifestations in different musical contexts remain subjects for future investigation.

\subsection{Dataset Preprocessing}

The \textbf{``Deutschl''} dataset of the \textbf{Essen Folk Song Collection} \citep{schaffrath_essen} consists of German folk melodies in \textbf{Kern format} (\texttt{.krn}) and all music related preprocessing tasks were carried out through a custom Python pipeline based on the \emph{music21} library \citep{cuthbert_music21}. 
    
The dataset was divided randomly by setting the seed parameter to 1 (\texttt{r\_state = 1}), splitting into an \textbf{80\% training set}, a \textbf{6\% validation set}, and a \textbf{14\% testing set} to guarantee reproducibility. All pieces of music were filtered systematically to discard those with notes or rests having durations not acceptable in the range \textbf{[0.25, 0.5, 0.75, 1, 1.5, 2, 3, 4] quarter lengths}. Each piece was further standardized with respect to the key by transposing all major-key pieces to \textbf{C major} and all minor-key pieces to \textbf{A minor}, thereby removing all unwanted tonal differences that would otherwise cause the model to learn melodic patterns from key characteristics. For encoding, musical elements were converted into \textbf{symbolic representations}: notes were represented using \textbf{MIDI pitch numbers}, rests using \texttt{`r'}, and duration using underscore signs \texttt{`\_'} for the subsequent time steps after the first symbolic representation. Each complete song, represented by a space-separated string of these symbols, was stored in separate text files before being merged under each dataset split, with songs separated by a delimiter made up of the symbol \texttt{`/'} repeated 64 times (\texttt{SEQUENCE\_LENGTH}), thus yielding three main files: \emph{train\_dataset}, \emph{val\_dataset}, and \emph{test\_dataset}.

In order to maintain consistent symbol interpretation across all model operations, \textbf{vocabulary mapping} was implemented. Symbol-to-index mappings were generated for each split, but only the mapping from the training split was used universally during preprocessing and training to avoid exposure to unseen symbols during evaluation. Input-target pairs for model training were constructed using a \textbf{sliding window approach} with \texttt{SEQUENCE\_LENGTH = 64}; an input consisted of 64 tokens in integer encoding and the target was the one token integer that directly followed it, with input vectors being \textbf{one-hot encoded} while target classes remained integers for optimization. \textbf{Seed sequences} were extracted to enable controlled melody generation and performance evaluation; hence 1,000 songs across all splits were randomly selected (500 from training, 100 from validation, and 400 from testing), and the first 20 characters were extracted from each encoded sequence. These standardized seed files were used as universal prompts in generative model testing, allowing for a reproducible evaluation of model performance across various experimental conditions while preserving enough diversity in the seed material to exhaustively test the generative capabilities of the model.

\subsection{Model Architecture}

This section presents the design and mathematical formulation of three neural network architectures implemented for \emph{symbolic music generation}: a \textbf{baseline LSTM model}, a \textbf{decoder-only Transformer model}, and a \textbf{hybrid Transformer-LSTM model}.
    
\subsubsection{Problem Formulation}

Let the symbolic melody be represented as a sequence of discrete tokens $\mathcal{X} = \{x_1, x_2, \ldots, x_T\}$, where each $x_t \in \mathcal{V}$ is a token from a finite vocabulary $\mathcal{V}$ consisting of musical symbols (MIDI pitches, rests, and duration markers). The objective is to train a model $\mathcal{M}_\theta$, parameterized by $\theta$, to learn the conditional distribution:

Given a seed sequence $x_{1:t}$, the model can \textbf{autoregressively generate} the continuation $\hat{x}_{t+1:T}$, where $\hat{x}_{t+1} = \arg\max P(x_{t+1} \mid x_{1:t}; \theta)$. The training objective minimizes the \textbf{negative log-likelihood} (Equation~\ref{eq:neg_log}) of the correct next token at each step:

\begin{equation}
\mathcal{L}(\theta) = -\sum_{t=1}^{T-1} \log P(x_{t+1} \mid x_1, x_2, \ldots, x_t; \theta)
\label{eq:neg_log}
\end{equation}

Each model variant approaches this optimization through different mechanisms for encoding $x_{1:t}$ and estimating $P(x_{t+1} \mid x_{1:t})$, as detailed in the following subsections.

\subsubsection{LSTM Baseline Model}

\begin{figure}[t]
    \centering
    \includegraphics[width=0.8\textwidth]{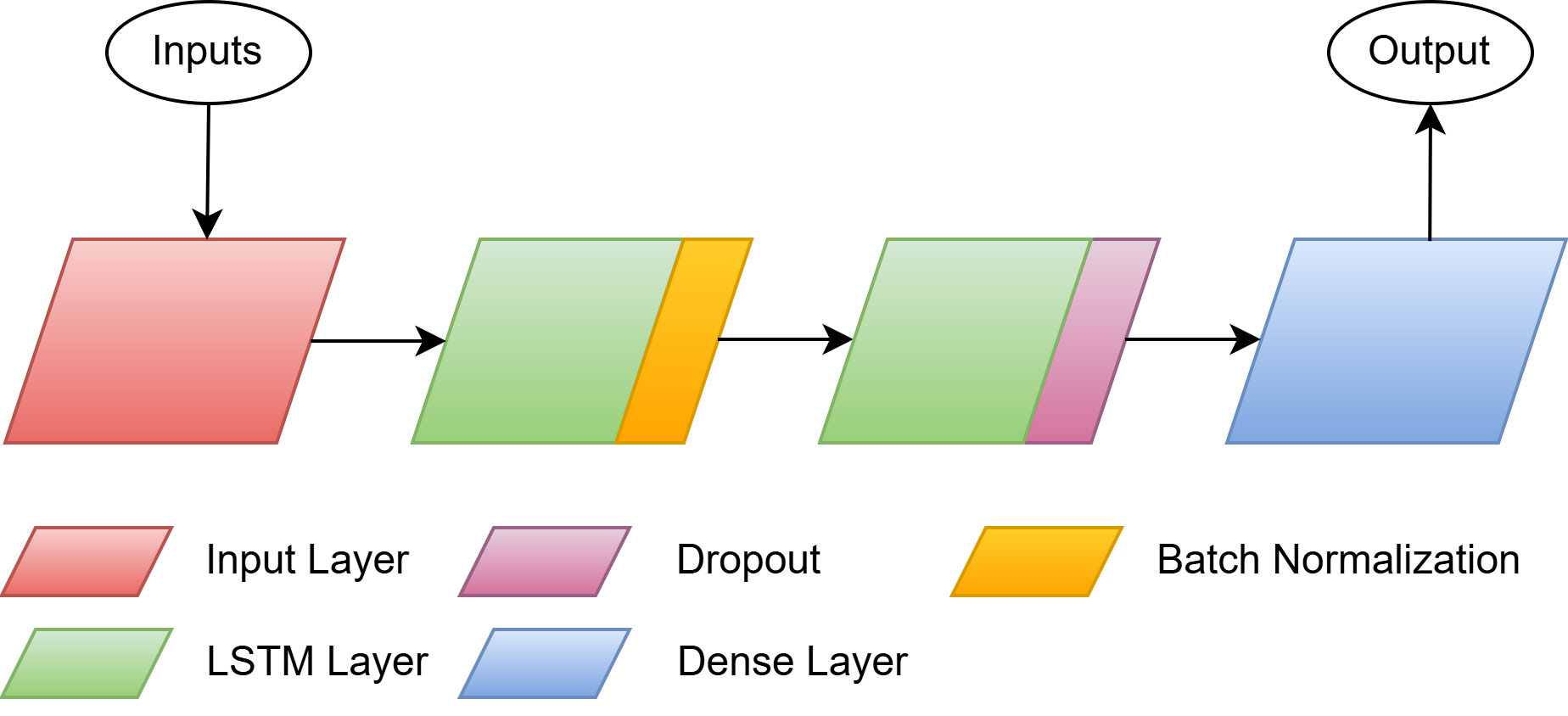}
    \caption{LSTM Architecture}
    \label{fig:lstm_architecture}
\end{figure}

The baseline model employs a two-layer \textbf{Long Short-Term Memory (LSTM)} network for \textbf{next-token prediction} in symbolic music sequences, processing fixed-length input sequences of 64 time steps with each time step represented as a one-hot encoded vector over a vocabulary of 45 tokens. The model consists of an \textbf{input layer} accepting sequences of shape $(64, 45)$, followed by the first LSTM layer with 256 units and \texttt{return\_sequences=True}, incorporating \textbf{L2 regularization} ($\lambda=0.01$) to penalize large weights. \textbf{Batch normalization} stabilizes training by normalizing layer inputs, while a second LSTM layer with 256 units outputs the final sequence representation. A \textbf{dropout layer} with rate 0.2 prevents overfitting by randomly zeroing neurons during training, and a \textbf{dense output layer} with softmax activation produces probability distributions over 45 tokens. The complete architecture is illustrated in Fig.~\ref{fig:lstm_architecture}.

Let $\mathcal{E}: \mathcal{V} \rightarrow \mathbb{R}^{d}$ be an \textbf{embedding function} mapping each token $x_t$ to a dense vector $e_t = \mathcal{E}(x_t)$. The LSTM processes this sequence as:

\begin{align}
    h_t &= \text{LSTM}(e_t, h_{t-1}; \theta_{\text{LSTM}}) \label{eq:lstm_hidden_state} \\
    \hat{y}_{t+1} &= \text{softmax}(W h_t + b) \label{eq:lstm_output}
\end{align}

where $h_t \in \mathbb{R}^d$ represents the hidden state at time $t$ (Equation~\ref{eq:lstm_hidden_state}), and $\hat{y}_{t+1} \in \mathbb{R}^{|\mathcal{V}|}$ is the \textbf{probability distribution over the next token} (Equation~\ref{eq:lstm_output}). Training employs \textbf{sparse categorical crossentropy loss} with \textbf{Adam optimizer} (learning rate = 0.001) and batch size of 64 for up to 50 epochs, incorporating \textbf{early stopping} (patience = 5), \textbf{model checkpointing} for best model retention, and \textbf{ReduceLROnPlateau} for learning rate adjustment during validation plateaus.

\subsubsection{Transformer Model}

\begin{figure}[t]
    \centering
    \includegraphics[width=0.9\textwidth]{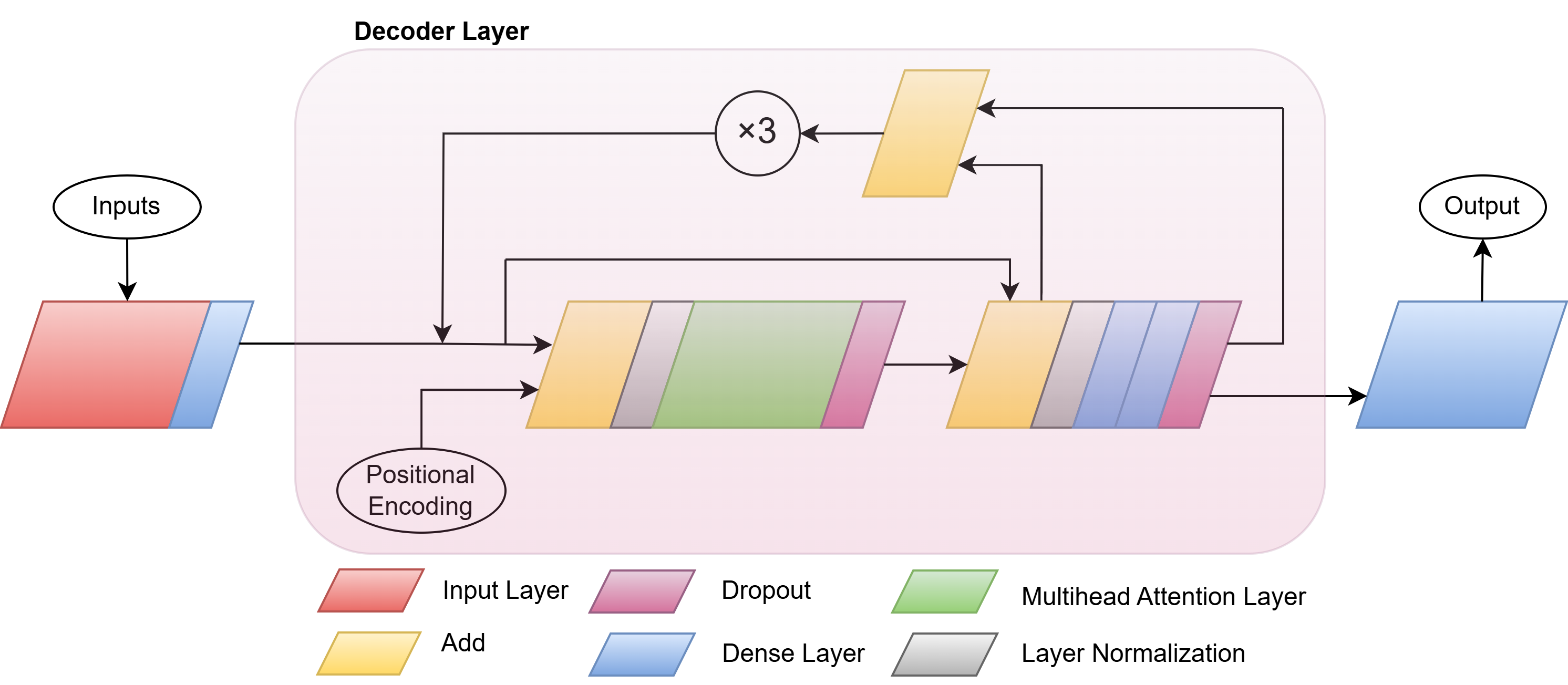}
    \caption{Transformer Architecture (Decoder-Only)}
    \label{fig:transformer_architecture}
\end{figure}

The Transformer model follows a \textbf{decoder-only architecture} inspired by \textbf{autoregressive language modeling}, designed to capture \textbf{long-range dependencies} and \textbf{structural patterns} in symbolic music through \textbf{self-attention mechanisms}. The model begins with an \textbf{input layer} accepting sequences of shape $(64, 45)$, followed by an \textbf{embedding layer} providing dense projection to $d_{\text{model}} = 256$-dimensional space. \textbf{Sinusoidal positional encodings} preserve sequence order information since attention mechanisms lack inherent position awareness. The core consists of a \textbf{4-layer transformer decoder stack}, where each layer includes \textbf{multi-head self-attention} (8 heads) that enables parallel attention to different sequence positions, \textbf{residual connections} that facilitate gradient flow, \textbf{layer normalization}, and a two-layer \textbf{feedforward network} ($d_{\text{ff}} = 1024$) with \textbf{dropout} (rate = 0.2). The output layer processes the final hidden state through \textbf{softmax} for next-token prediction, as shown in Fig. \ref{fig:transformer_architecture}.

The Transformer processes the input sequence $x_{1:t}$ holistically using self-attention:
\begin{align}
    e_t &= \mathcal{E}(x_t) + \mathcal{P}(t) \label{eq:transformer_embed} \\
    Z_t &= \text{TransformerDecoder}(e_{1:t}; \theta_{\text{Tr}}) \label{eq:transformer_decoder} \\
    \hat{y}_{t+1} &= \text{softmax}(W Z_t + b) \label{eq:transformer_output}
\end{align}

where $\mathcal{P}(t) \in \mathbb{R}^d$ represents \textbf{positional encoding} (Equation~\ref{eq:transformer_embed}), $Z_t$ is the output of the final decoder layer at timestep $t$ (Equation~\ref{eq:transformer_decoder}), and $\hat{y}_{t+1}$ is the predicted probability distribution (Equation~\ref{eq:transformer_output}). The model employs \textbf{sparse categorical crossentropy loss} with \textbf{Adam optimizer} using an \textbf{inverse square root learning rate scheduler} \emph{as originally proposed in} \citep{vaswani2017attention} that gradually reduces learning rate after 4000 \textbf{warm-up steps}. \textbf{JIT compilation} enhances performance optimization, while training includes \textbf{early stopping} (patience = 5), \textbf{model checkpointing}, and \textbf{learning rate logging callbacks}.

\subsubsection{Transformer-LSTM Hybrid Model}

\begin{figure}[t]
    \centering
    \includegraphics[width=0.9\textwidth]{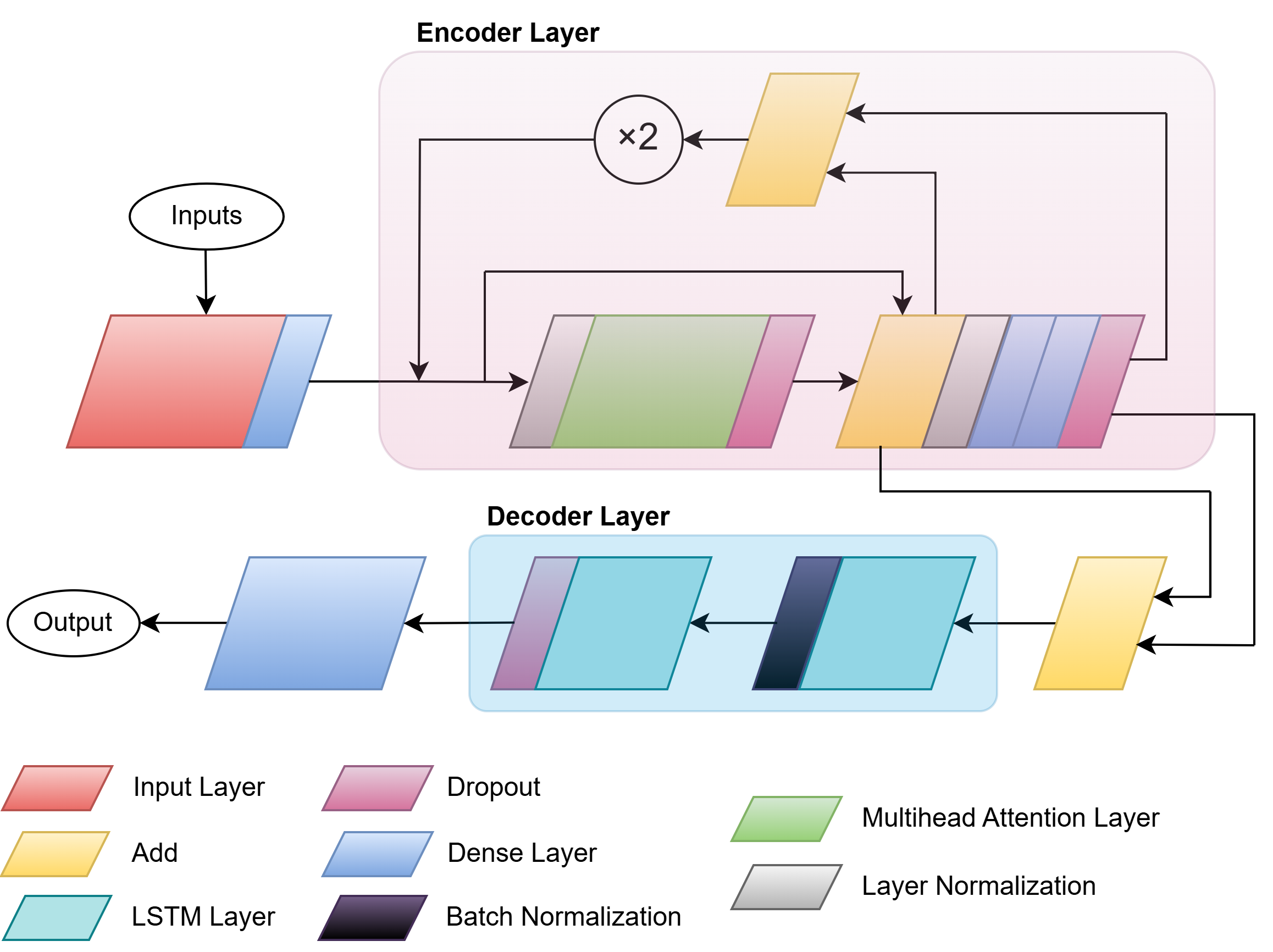} 
    \caption{Transformer-LSTM Hybrid Architecture (Encoder-Decoder)}
    \label{fig:hybrid_architecture}
\end{figure}

The \textbf{hybrid architecture} integrates a \textbf{Transformer encoder} with an \textbf{LSTM decoder} to leverage both \textit{global attention mechanisms} and \textit{sequential modeling capabilities}, designed to capture \textit{structural motifs} and \textit{temporal progression} more effectively. The model processes one-hot encoded sequences of shape $(64, 45)$ through an \textbf{embedding layer} projecting to 256-dimensional latent space. A \textbf{3-layer Transformer encoder stack} follows, where each layer incorporates \textbf{layer normalization} ($\epsilon = 1 \times 10^{-6}$), \textbf{multi-head self-attention} (8 heads) for capturing global dependencies, \textbf{position-wise feedforward networks} ($d_{\text{ff}} = 512$), \textbf{residual connections}, and \textbf{dropout} (rate = 0.1). The LSTM decoder consists of two layers: the first with 256 units, \texttt{return\_sequences=True}, and \textbf{L2 regularization} ($\lambda=0.01$), followed by \textbf{batch normalization} and a second LSTM layer with 256 units and \textbf{dropout} (rate = 0.2). A \textbf{dense output layer} with softmax activation produces \textbf{probability distributions} over 45 tokens, with the complete hybrid architecture depicted in Fig.~\ref{fig:hybrid_architecture}.

In this hybrid approach, input tokens are first embedded (Equation~\ref{eq:hybrid_embed}) and encoded globally using the Transformer encoder (Equation~\ref{eq:hybrid_encoder}):

\begin{align}
    e_t &= \mathcal{E}(x_t) \label{eq:hybrid_embed} \\
    \tilde{e}_t &= \text{TransformerEncoder}(e_{1:t}; \theta_{\text{Enc}}) \label{eq:hybrid_encoder}
\end{align}

The encoded sequence $\tilde{e}_t$ is subsequently processed by the LSTM decoder (Equations~\ref{eq:decoder_lstm} –~\ref{eq:decoder_output}) to model temporal dependencies:

\begin{align}
    h_t &= \text{LSTM}(\tilde{e}_t, h_{t-1}; \theta_{\text{Dec}}) \label{eq:decoder_lstm} \\
    \hat{y}_{t+1} &= \text{softmax}(W h_t + b) \label{eq:decoder_output}
\end{align}

This formulation combines \textbf{global context modeling} from the Transformer encoder with \textbf{fine-grained temporal modeling} from the LSTM decoder, aiming to achieve superior performance in symbolic music synthesis. The hybrid model uses \textbf{sparse categorical crossentropy loss} with \textbf{Adam optimizer} (learning rate = 0.001) and batch size of 64 for up to 50 epochs, incorporating \textbf{early stopping} (patience = 5), \textbf{model checkpointing}, and \textbf{ReduceLROnPlateau} callbacks (patience = 2, factor = 0.2) for adaptive learning rate adjustment.

\subsection{Training Procedure}
The training of all models was carried out in a supervised manner on \textbf{time-series datasets} of \textbf{symbolic music sequences} derived from the Kern dataset of German Folklore melodies, where \texttt{.krn} files were converted to \textbf{trainable time-series representation}. The input sequence refers to previous notes, and the output is the next predicted note. We split the dataset into training, validation, and test sets with an \textbf{80:6:14} ratio using \texttt{random\_state=1} for balancing classes, where each file represents one melody and content was concatenated under respective sets into single files with 64 \texttt{'/'} as delimiter. Input sequences were \textbf{one-hot encoded} using mappings created for the training set and applied globally across validation and test sets, with targets represented as integer class labels. The \textbf{Adam optimizer} was used across all model variants for efficient, adaptive gradient descent \citep{kingma2014adam}. However, the Transformer architecture applied a \textbf{custom learning rate scheduler} based on an \textbf{inverse square root decay} strategy, adjusting the LR as a function of \textbf{model dimension} and \textbf{training step} with an additional \textbf{warm-up period} to smooth out changes during the early updates \citep{vaswani2017attention}. For the loss function, \textbf{sparse categorical cross-entropy} was used by all models, which is suitable for multiclass classification with sparse, integer-labeled targets, computing the \textbf{negative log-probability} of the correct class while encouraging the correct next note to be assigned a high confidence value. During training, \textbf{early stopping} was applied with a patience of 5 evaluation steps on the validation loss \citep{li2019gradient}, with automatic restoration of the best model weights and checkpointing of the model at its least validation loss. Training was capped at 50 epochs, with a batch size of 64, maintaining detailed logs for further examination and visualization.

To assess generative performance, a structured pipeline was implemented beginning with seed-based melody generation from \textbf{1,000 pre-generated seed sequences} created by taking the first 6 notes from random files across train (500), validation (100), and test (400) sets, serving as initial prompts for \textbf{autoregressive generation}. The generation process involved tokenization through one-hot encoding of each musical event (pitch, rest, duration) into structured discrete representations, followed by \textbf{iterative sequence generation} where the model received the current sequence, predicted the next event as an integer index, appended this prediction to the current sequence, and passed the updated sequence back for the next prediction step.

This process continued for a fixed number of time steps, simulating melody continuation beyond the initial seed, with all models employing \textbf{temperature sampling} at $\tau = 0.7$ to introduce controlled stochasticity in the generation process. Temperature sampling modifies the probability distribution over the vocabulary by rescaling logits before applying softmax: 

\[
P(x_{t+1} \mid x_{1:t}) = \mathrm{softmax}\left(\frac{\text{logits}}{\tau}\right),
\]

where $\tau < 1$ sharpens the distribution (more deterministic) and $\tau > 1$ flattens it (more random) \citep{wang2020contextual}. This approach enables the model to \textit{explore multiple plausible continuations} rather than always selecting the most probable next token, thereby capturing the creative variability inherent in musical composition while maintaining structural coherence. The generated melodies were subsequently evaluated using \textbf{11 local variance measures} and \textbf{5 global variance measures}, with comprehensive evaluation metrics stored and summarized for analysis.

\subsection{Quantitative Evaluation}
The selection of \textbf{17 evaluation metrics}, described in Table~\ref{tab:local-metrics-desc}  and Table~\ref{tab:global-metrics-desc}, was based on established frameworks for computational music analysis, encompassing both local (micro-level) and global (macro-level) musical characteristics. Local variance metrics capture \textbf{immediate melodic behaviors} such as pitch transitions and rhythmic patterns, reflecting \textit{perceptual elements} identified in music cognition research \citep{huron2008sweet}. Global variance metrics assess \textbf{structural properties} including entropy-based measures of \textit{predictability and diversity indices}, following approaches established in computational creativity assessment \citep{pearce2006expectation}. This comprehensive metric set enables multidimensional evaluation of generative quality while avoiding reliance on single measures that may not capture the complexity of musical structure.

The 17 metrics employed are grounded in established music-theoretical principles and computational musicology research. Local variance metrics assess immediate melodic behaviors such as pitch relationships and rhythmic patterns, drawing from contour theory \citep{morris1993new}, species counterpoint \citep{fux1965study}, and motivic analysis \citep{schoenberg2006musical}. Global variance metrics evaluate structural properties including entropy-based measures following information-theoretic approaches to musical complexity \citep{huron2008sweet}, and harmonic complexity measures based on tonal pitch space theory \citep{lerdahl2001tonal}. This multidimensional approach enables comprehensive assessment while maintaining theoretical grounding in established musicological frameworks.

\begin{table}[t]
    \centering
    \caption{Local Variance Metrics}
    \begin{tabular}{@{}p{4cm} p{8.5cm} p{3cm}@{}}
    \toprule
    \textbf{Metric (with Theoretical Foundation)} & \textbf{Description} & \textbf{Range} \\
    \midrule
    Pitch Variance (Melodic contour analysis \citep{morris1993new}) & Measures statistical variability between adjacent pitches, quantifying melodic smoothness versus angularity & 0 (no variance) to high \\[6pt]\hline
    
    Pitch Range (Registral analysis theory \citep{morris1993new}) & Distance between highest and lowest pitches, representing the melodic ambitus in traditional music theory & 0 to maximum interval \\[6pt]\hline
    
    Rhythmic Variance (Rhythmic complexity theory \citep{london2012hearing}) & Statistical variation in note duration values, reflecting temporal complexity & 0 (uniform) to high \\[6pt]\hline
    
    Note Density (Temporal density analysis \citep{london2012hearing}) & Number of note onsets per temporal unit, corresponding to surface rhythm density & 0 to high \\[6pt]\hline
    
    Rest Ratio (Phrase structure analysis \citep{caplin1998classical}) & Proportion of silence relative to sounding events, relating to phrase structure theory & 0 (no rests) to 1 (only rests) \\[6pt]\hline
    
    Interval Variability (Species counterpoint theory \citep{fux1965study}) & Average fluctuation in melodic intervals, measuring step-wise versus leap-wise motion & 0 (constant) to high \\[6pt]\hline
    
    Note Repetition (Motivic analysis theory \citep{schoenberg2006musical}) & Frequency of pitch repetition, inversely related to melodic diversity & 0 (no repetition) to 1 (all repeated) \\[6pt]\hline
    
    Contour Stability (Contour theory \citep{morris1993new}) & Consistency of melodic direction changes, based on contour reduction theory & 0 (unstable) to 1 (stable) \\[6pt]\hline
    
    Syncopation (Metric theory \citep{longuet1982perception}) & Quantification of off-beat note placements following established syncopation models & 0 (none) to high \\[6pt]\hline
    
    Harmonic Tension (Tonal harmony theory \citep{piston1959harmony}) & Level of harmonic dissonance based on consonance-dissonance gradations & 0 (no tension) to high \\[6pt]\hline
    
    KL Divergence (Information-theoretic music analysis \citep{huron2008sweet}) & Divergence from reference pitch distribution, measuring stylistic deviation & 0 (identical) to high \\
    \bottomrule
    \end{tabular}

    \label{tab:local-metrics-desc}
\end{table}

\begin{table}[t]
    \centering
    \caption{Global Variance Metrics}
    \begin{tabular}{@{}p{4cm} p{8.5cm} p{3cm}@{}}
    \toprule
    \textbf{Metric (with Theoretical Foundation)} & \textbf{Description} & \textbf{Range} \\
    \midrule
    Pitch Entropy (Music expectation theory \citep{huron2008sweet}) & Shannon entropy of pitch class distribution, quantifying pitch unpredictability & 0 (predictable) to $\log_2(N)$ \\[6pt]\hline
    
    Rhythmic Entropy (Music expectation theory \citep{huron2008sweet}) & Shannon entropy applied to rhythm patterns, measuring temporal unpredictability & 0 (predictable) to $\log_2(N)$ \\[6pt]\hline
    
    Motif Diversity Index (Motivic analysis theory \citep{schoenberg2006musical}) & Variety of recurring melodic patterns, based on motivic segmentation analysis & 0 (single motif) to high \\[6pt]\hline
    
    Harmonic Complexity (Tonal pitch space theory \citep{lerdahl2001tonal}) & Richness of implied harmonic structures derived from tonal pitch space theory & 0 (simple) to high \\[6pt]\hline
    
    Contour Variability (Contour analysis theory \citep{morris1993new}) & Statistical variance in melodic shape patterns extending contour analysis & 0 (similar) to high \\[6pt]\hline
    
    Tonal Drift (Key analysis theory \citep{krumhansl2001cognitive}) & Shift in tonal center over time, measuring key stability using key-finding principles & 0 (stable) to high \\
    \bottomrule
    \end{tabular}
    \label{tab:global-metrics-desc}
\end{table}

Once \textbf{1,000 melodies} were generated per model, they were passed through a pipeline to evaluate the \textbf{comprehensive set of 17 musical quality metrics}. For each metric, the following \textbf{statistical summaries} were computed across all 1,000 generated melodies: \textbf{SUM}, \textbf{MIN}, \textbf{MAX}, \textbf{MEAN}, \textbf{MEDIAN}, and \textbf{STANDARD DEVIATION (STD)}. These aggregated values allowed for both \textbf{individual and comparative evaluation} of the models' generative capabilities \citep{ji2023survey}.

\textbf{Architectural ablation studies} involving \textbf{hybrid variants} were put in structure to deepen the understanding of the effect of each architectural component on melody generation. The studies were conducted for four main categories: the \textbf{encoder design} varying in the number of \textit{Transformer layers} and \textit{attention heads} so as to affect the modeling of the global structure \citep{zou2021melons}; changes to the \textbf{decoder structure} concerning different numbers of \textit{LSTM layers} and other \textit{hidden unit sizes} for the study of the temporal model sensitivity; the \textbf{input and positional representation} considering the \textit{absence of embedding layers} and \textit{presence of positional encodings}; and \textbf{regularization strategies} for different \textit{dropout configurations} and the \textit{strength of L2 regularization}, studying those features with respect to \textit{model robustness} and control of \textit{overfitting}. Various parameters for the architectural study included those referenced throughout the results \textbf{A1a} (fewer Transformer layers), \textbf{B3a} (lower LSTM capacity), or \textbf{D1f} (high L2 penalty), which were not designed to lay emphasis on achieving one best performance but rather on gathering \textbf{architectural interpretability} in order to discuss \textbf{tradeoffs between expressiveness, stability, and musical coherence} in the results generated.

\subsection{Human Perceptual Evaluation}
To validate the computational metrics and address the critical gap between algorithmic assessment and human perception, a human evaluation study was conducted using a \textbf{structured perceptual assessment framework}. Three representative melody samples from each model architecture (LSTM, Transformer, and Hybrid) were selected based on their proximity to the mean performance across computational metrics, ensuring representative rather than cherry-picked examples.

A total of \textbf{21 participants} (ages 18-44, with 23.8\% having formal musical training, 33.8\% amateur experience, and 42.4\% no training) evaluated each of the \textbf{9 musical clips} through a structured online survey. Participants used \textbf{varied listening setups} including headphones (61.9\%), laptop speakers (28.6\%), and external speakers (9.5\%) to capture realistic listening conditions. Each clip was assessed across five perceptual dimensions using 10-point Likert scales: \textbf{musicality} (pleasantness and musical quality), \textbf{coherence} (logical melody structure), \textbf{originality} (melodic interest and novelty), \textbf{naturalness} (human-like quality), and \textbf{overall rating}. The evaluation order was randomized to minimize ordering effects, and participants provided informed consent for voluntary participation.

\section{Results and Discussions} \label{section:results}

\subsection{Experimental Setup}

The \textbf{Deutschl} dataset of the \textbf{
Essen Folk Song Collection} \citep{schaffrath_essen} has been used for running the experiments. This framework, being complete, featured \textbf{symbolical representations} of German folk melodies in .krn format. This dataset was shortlisted for training and testing generative music models on account of its common notation and \textbf{extremely rich melodic variation}. For establishing robust and statistically significant evidence, 1,000 melodies were generated for each of the three model architectures under study: \textbf{LSTM}, \textbf{Transformer}, and \textbf{Transformer-LSTM Hybrid}. Generations across all models were made consistent by using a single set of \textbf{randomly sampled seeds}, so that results were not potentially biased by using different initiation conditions and also for a fair comparative analysis of the architectures. Every one of the 3000 generated melodies was subjected to an evaluation process that looked at \textbf{17 musical quality metrics}, which covered both structural and aesthetic considerations in music composition. This varied statistical approach provided a thorough characterization of the model performances concerning both the \textbf{central tendencies} and variability in \textbf{musicality and structural diversity} across model variants.

All computational procedures, including model training and inference operations, were executed on the \emph{RunPod cloud platform} using a high-performance configuration consisting of an \emph{AMD EPYC 7702 64-Core Processor} with \emph{47 GB of system memory} and an \emph{NVIDIA RTX 4000 Ada GPU} featuring \emph{20 GB of VRAM}. GPU acceleration was leveraged throughout all experimental phases, enabling efficient large-scale training and generation processes essential for comprehensive model evaluation.

\subsection{Comparative Performance Analysis}

\subsubsection{Local Variance Metrics Performance}

The comparative analysis of local variance metrics revealed distinct performance characteristics across the three model architectures, as summarized in Table \ref{tab:local_variance_metrics}. The Transformer-LSTM Hybrid model demonstrated superior performance across most local metrics, generating melodies with significantly \textbf{enhanced expressiveness} and \textbf{diversity} in both pitch and rhythm dimensions. Specifically, the hybrid model achieved a pitch variance of 7.11, substantially higher than both the LSTM (3.57) and Transformer (3.63) baselines, indicating \textbf{greater melodic expressiveness}. Similarly, the hybrid model exhibited superior interval variability (3.94) compared to the identical performance of LSTM and Transformer models (1.69), suggesting more \textbf{sophisticated harmonic progression} capabilities. The rhythmic variance results further reinforced this pattern, with the hybrid model achieving 0.25 compared to 0.12 and 0.10 for LSTM and Transformer respectively, demonstrating \textbf{enhanced rhythmic complexity} while \textbf{maintaining structural coherence}. The KL divergence was also highest for the hybrid model (1.05), compared to LSTM (0.83) and Transformer (0.77), indicating \textbf{greater creative deviation} from the training distribution while \textbf{maintaining melodic coherence}.

\begin{table}[t]
\centering
\caption{Local Variance Metrics Summary}
\begin{tabular}{lccc}
\hline
\textbf{Metric} & \textbf{LSTM} & \textbf{Transformer} & \textbf{Hybrid} \\
\hline
Pitch Variance       & 3.57 & 3.63 & 7.11 \\
Pitch Range          & 4.24 & 4.33 & 5.21 \\
Rhythmic Variance    & 0.12 & 0.10 & 0.25 \\
Note Density         & 1.08 & 1.14 & 0.92 \\
Rest Ratio           & 0.07 & 0.09 & 0.06 \\
Interval Variability & 1.69 & 1.69 & 3.94 \\
Note Repetition      & 1.78 & 1.90 & 1.45 \\
Contour Stability    & 0.58 & 0.65 & 0.78 \\
Syncopation          & 1.54 & 1.65 & 1.27 \\
Harmonic Tension     & 0.44 & 0.44 & 0.38 \\
KL Divergence        & 0.83 & 0.77 & 1.05 \\
\hline
\end{tabular}
\label{tab:local_variance_metrics}
\end{table}

\subsubsection{Global Variance Metrics and Structural Complexity}

The global variance metrics, presented in Table \ref{tab:global_variance_metrics}, reinforced the Transformer-LSTM Hybrid model's superiority in producing \textbf{musically rich} and \textbf{structurally complex} melodies. Pitch entropy and rhythmic entropy values supported this observation, with the hybrid model achieving 2.70 and 1.94 respectively, significantly outperforming both baseline models. The Motif Diversity Index results were particularly noteworthy, with the hybrid model scoring 0.86 compared to 0.76 for LSTM and 0.74 for Transformer, demonstrating \textbf{superior capability} in generating \textbf{varied and complex melodic motifs}. Additionally, the tonal drift metric showed the hybrid model's enhanced ability to create \textbf{dynamic melodic progression} (6.41) while the LSTM model remained more \textbf{conservative} (4.30) and the Transformer showed \textbf{intermediate behavior} (3.58).

\begin{table}[t]
\centering
\caption{Global Variance Metrics Summary}
\begin{tabular}{lccc}
\hline
\textbf{Metric} & \textbf{LSTM} & \textbf{Transformer} & \textbf{Hybrid} \\
\hline
Pitch Entropy           &  2.53  &  2.43  &  2.70  \\
Rhythmic Entropy        &  1.33  &  1.20  &  1.94  \\
Motif Diversity Index   &  0.76  &  0.74  &  0.86  \\
Harmonic Complexity     &  0.52  &  0.51  &  0.90  \\
Contour Variability     &  23.61 &  24.16 &  45.51 \\
Tonal Drift             &  4.30  &  3.58  &  6.41  \\
\hline
\end{tabular}
\label{tab:global_variance_metrics}
\end{table}

\subsubsection{Comparative Model Performance Characteristics}

The comprehensive analysis across all 17 evaluation metrics revealed three distinct performance profiles corresponding to each model architecture. The LSTM model demonstrated the strongest alignment with the training distribution, producing \textbf{conservative and genre-faithful melodies} that closely adhered to traditional German folk music characteristics. This model excelled in maintaining harmonic tension (0.44), rest ratio (0.070), and note repetition patterns (1.78) that are characteristic of structured folk melodies. The Transformer model exhibited \textbf{balanced performance} with moderate variability across most metrics, positioning itself as an intermediate architecture between the conservative LSTM and the more expressive hybrid model. While the Transformer showed slight improvements in note density (1.14) and syncopation (1.65) compared to LSTM, it maintained relatively stable performance across global metrics, suggesting limited capacity for long-range structural innovation. The hybrid Transformer-LSTM architecture, however, delivered the most \textbf{expressive and structurally rich melodies}, exhibiting the highest pitch variance (7.11), interval variability (3.94), and motif diversity (0.86), along with improved contour stability (0.78) and rhythmic entropy (1.94). Despite a slight drop in rest ratio (0.06) and harmonic tension (0.38), it \emph{outperformed both baselines in terms of melodic complexity and global structure}, as evidenced by its superior contour variability (45.51) and harmonic complexity (0.90), confirming it as the \textbf{most musically balanced} model among the three.

The relationship between predictive accuracy and musical quality requires careful interpretation in generative contexts. While the hybrid model achieved slightly lower test accuracy compared to baseline models, this \textbf{reduction corresponds to increased generative diversity rather than modeling failure}. Higher accuracy typically indicates closer adherence to training data patterns, which may constrain creative exploration. The hybrid architecture's marginally reduced accuracy (77.56\% vs. baseline values) enables \textbf{greater melodic variability} while maintaining sufficient structural coherence, as evidenced by improved performance across diversity and expressiveness metrics.

\subsubsection{Qualitative Analysis and Symbolic Pattern Recognition}

Beyond quantitative metrics, qualitative examination of generated token sequences and symbolic patterns revealed distinctive compositional characteristics for each model architecture. The LSTM-generated melodies consistently exhibited \textbf{short, repetitive note sequences} and \textbf{stable contour patterns}, aligning with the structured phrasing and motif reuse typical in German folk melodies. Visual inspection of the generated sequences confirmed the model's tendency toward \textbf{conservative melodic construction} with predictable phrase structures. Transformer outputs displayed notably \textbf{more variation} in rhythm and pitch transitions, occasionally introducing \emph{abrupt changes} or \emph{denser note clusters} that enhanced diversity but sometimes resulted in \textbf{irregular phrasing} or \textbf{melodic drift}. The Transformer-LSTM Hybrid model demonstrated the most \emph{balanced approach}, effectively combining repetition and variation through \textbf{localized motif recurrence} and \textbf{global contour evolution}, supporting the quantitative findings regarding its superior performance across both micro- and macro-level musical features.

\subsubsection{Integrated Performance Assessment and Model Suitability}

The comprehensive evaluation across local and global metrics, supported by qualitative analysis and human perception validation, established clear performance hierarchies and application domains for each model architecture. The LSTM model emerged as the optimal choice for applications requiring \textbf{strict stylistic consistency} and \textbf{genre fidelity}, making it particularly suitable for \emph{traditional music preservation} and educational applications. The Transformer model's balanced performance profile positioned it as an effective middle-ground architecture, offering \textbf{moderate creative flexibility} while maintaining reasonable adherence to training data characteristics. The Transformer-LSTM Hybrid model conclusively outperformed both baselines in \textbf{expressiveness}, \textbf{motif richness}, and \textbf{structural diversity}, establishing it as the superior architecture for generative tasks requiring the \emph{optimal balance between fidelity and novelty}. Thus, The hybrid model demonstrated \textbf{enhanced generative diversity} through \textit{increased variance metrics while maintaining structural coherence}, indicating improved capacity for creative exploration within \textbf{musically valid parameter spaces}. This increased variability, as evidenced by higher KL divergence and entropy measures, reflects the model's ability to generate \textit{novel melodic patterns} that \textit{extend beyond direct replication of training examples} while preserving essential musical characteristics.

\subsection{Ablation analysis of the Hybrid Model}

\subsubsection{Comprehensive Architectural Component Analysis}

To systematically investigate the influence of individual architectural components within the \textbf{hybrid Transformer-LSTM model}, a comprehensive series of \textbf{controlled ablation experiments} was conducted. Each experimental variant involved a precisely targeted modification to either the \textbf{encoder structure}, \textbf{decoder configuration}, \textbf{input processing mechanism}, or \textbf{regularization parameters}, while maintaining all other components constant. 

This methodical approach enabled the \textbf{isolation of specific architectural contributions} to overall model performance. Performance assessment was conducted using the identical \textbf{17 evaluation metrics} described in the methodology section, ensuring consistent and interpretable comparisons across all variants.

The experimental design encompassed a comprehensive range of architectural modifications, including variations in \textbf{encoder depth} (\textit{A1a}: reducing Transformer layers from 3 to 1), \textbf{attention mechanisms} (\textit{A2a}: decreasing attention heads from 8 to 4), \textbf{decoder structure} (\textit{B1}: simplifying from 2 to 1 LSTM layers; \textit{B3a}: reducing LSTM units from 256 to 128), \textbf{input processing strategies} (\textit{C1}: removing the embedding layer; \textit{C2}: adding positional encoding), and \textbf{regularization approaches} (\textit{A3}: complete dropout removal; \textit{D1a--D1f}: systematic L2 regularization variations across different penalty strengths).

\subsubsection{Encoder Architecture Modifications and Performance Impact}

The analysis of \textbf{encoder-focused modifications} revealed significant insights into the role of Transformer components in melodic generation quality. The \emph{shallow encoder variant} (\textbf{A1a}), which reduced Transformer layers from 3 to 1, demonstrated improved computational efficiency with a \textbf{58\% reduction in total parameters} (3,442,477 vs 8,177,453) and enhanced \textbf{test accuracy} (+0.62\%), indicating better generalization capabilities. However, this architectural simplification resulted in decreased \textbf{musical expressiveness}, with \textbf{pitch variance} dropping from 7.11 to 4.24 and \textbf{interval variability} reducing from 3.94 to 1.69, suggesting diminished \textbf{melodic diversity} and \textbf{harmonic sophistication}.

The \emph{reduced attention heads variant} (\textbf{A2a}), decreasing from 8 to 4 heads, exhibited similar patterns with improved \textbf{accuracy} (+0.90\%) but reduced \textbf{rhythmic complexity}, as evidenced by decreased \textbf{rhythmic variance} (0.10 vs 0.25) and \textbf{syncopation} (1.27 vs 1.65). Conversely, global metrics such as \textbf{KL divergence} and \textbf{entropy measures} showed mixed results, with the attention reduction variant demonstrating increased \textbf{pitch entropy} (2.70 vs 2.43) but decreased \textbf{rhythmic entropy} (1.20 vs 1.94), indicating complex trade-offs between local and global musical coherence.

\subsubsection{Decoder Structure Analysis and Sequential Memory Impact}

\textbf{Decoder-focused modifications} provided crucial insights into the role of LSTM components in maintaining \textbf{sequential coherence} and \textbf{temporal dependencies} within generated melodies. The \emph{simplified decoder variant} (\textbf{B1}), reducing LSTM layers from 2 to 1, maintained similar overall accuracy (77.37\% vs 77.56\%) but required significantly more \textbf{epochs to converge} (28 vs 2), indicating reduced learning efficiency. Musical analysis revealed substantial impacts on melodic characteristics, with \textbf{pitch variance} decreasing dramatically and \textbf{note density} dropping sharply, resulting in \textbf{sparser musical textures}. Interestingly, this modification led to increased \textbf{rhythmic variance} and \textbf{harmonic complexity}, suggesting that reduced sequential memory capacity forced the model to rely more heavily on local variation rather than long-term structural coherence.

The \emph{LSTM unit reduction variant} (\textbf{B3a}), decreasing from 256 to 128 units, demonstrated superior \textbf{generalization} with validation accuracy surpassing training accuracy, indicating reduced \textbf{overfitting tendencies}. This modification resulted in increased \textbf{musical diversity} through higher \textbf{entropy} and \textbf{motif diversity} metrics, while maintaining \textbf{computational efficiency} with a 9\% parameter reduction. These results demonstrate that \textbf{decoder capacity modifications} create complex trade-offs between \textbf{computational efficiency}, \textbf{convergence speed}, and \textbf{musical expressiveness}.

\begin{figure}[t]
\centering
\includegraphics[width=0.9\textwidth]{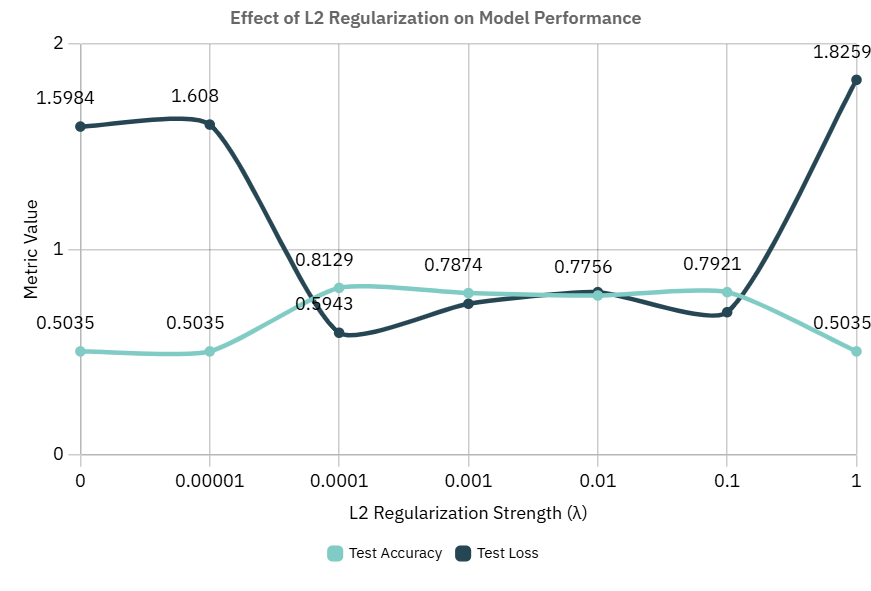} 
\caption{Impact of Ridge Regression (L2) on Predictive Accuracy}
\label{fig:l2_trend}
\end{figure}

\subsubsection{Input Processing and Regularization Strategy Evaluation}

\textbf{Input processing modifications} revealed the critical importance of \textbf{embedding layers} and \textbf{positional information} in musical sequence modeling. The \emph{embedding layer removal variant} (\textbf{C1}) achieved the highest test accuracy improvement (+7.05\%) and a dramatic \textbf{parameter reduction} (85\% fewer parameters), but at the cost of significantly reduced \textbf{musical expressiveness}, with lower \textbf{pitch} and \textbf{rhythmic variance} indicating flatter, less dynamic musical outputs. This finding underscores the embedding layer’s crucial role in capturing \textbf{sequence richness} and \textbf{melodic complexity}. The \emph{positional encoding addition variant} (\textbf{C2}) demonstrated enhanced \textbf{sequence-awareness} with improved accuracy and reduced loss, but required substantially more \textbf{epochs to converge} (46 vs 2), indicating the complexity of incorporating positional information effectively.

\textbf{Regularization analysis} through \emph{dropout removal} (\textbf{A3}) and \emph{L2 regularization variations} (\textbf{D1a--D1f}) provided definitive evidence for the necessity of proper \textbf{regularization} in maintaining \textbf{model stability} and \textbf{musical quality}. The complete removal of regularization components resulted in \textbf{catastrophic performance degradation}, with test accuracy dropping by 26.7\% and \textbf{musical diversity metrics} showing erratic, unstable behavior. These results confirm that regularization is essential for both \textbf{generalization} and \textbf{musical coherence}.

\begin{figure}[t]
\centering
\includegraphics[width=0.9\textwidth]{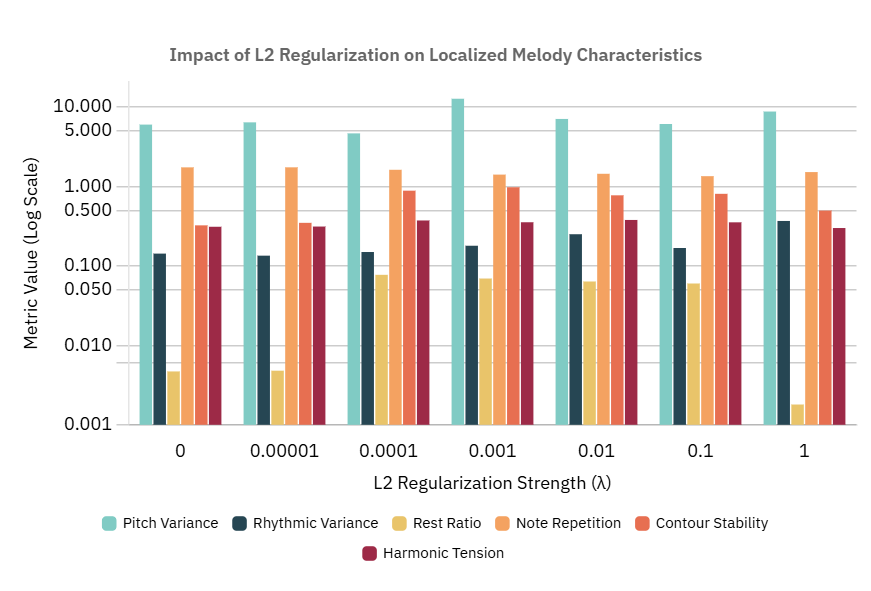} 
\caption{Impact of Ridge Regression (L2) on Local Variance}
\label{fig:l2_on_local}
\end{figure}

\begin{figure}[t]
\centering
\includegraphics[width=0.9\textwidth]{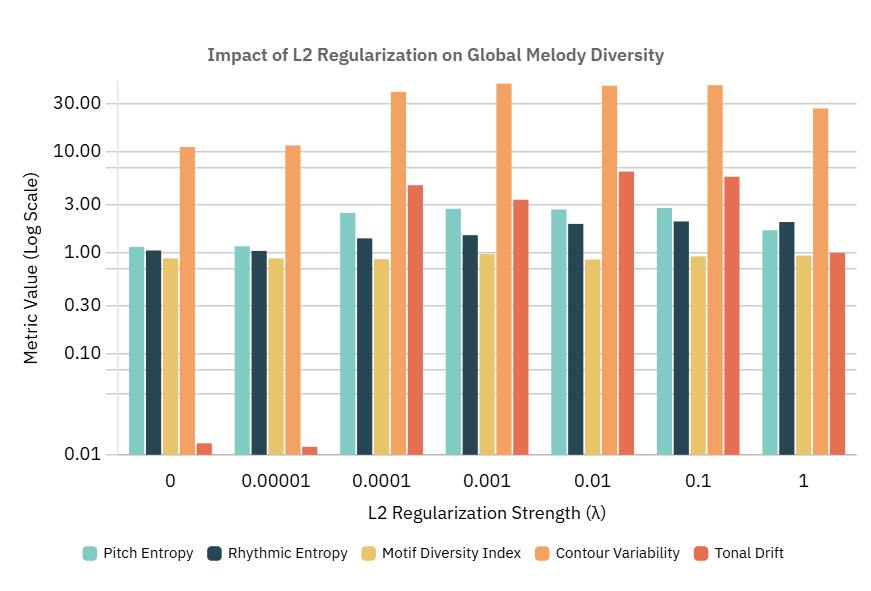} 
\caption{Impact of Ridge Regression (L2) on Global Variance}
\label{fig:l2_on_global}
\end{figure}

\subsubsection{L2 Regularization Optimization and Performance Curves}

The systematic analysis of \textbf{L2 regularization strength variations} (D1a--D1f) revealed a clear \textbf{optimal range} for balancing model flexibility and structural coherence in musical generation. The regularization strength sweep, encompassing values from 0 to 1.0, demonstrated a characteristic \textbf{performance curve} as comprehensively visualized across all three figures: Fig.~\ref{fig:l2_trend} illustrates the overall model performance trends, Fig.~\ref{fig:l2_on_local} shows local variance impacts, and Fig.~\ref{fig:l2_on_global} demonstrates global variance patterns across different regularization strengths.

Optimal performance was achieved with \textbf{moderate regularization values} (\( \lambda = 0.0001 \) to \( 0.001 \)), where the model attained the \textbf{highest test accuracy} (+3.73\% for \( \lambda = 0.0001 \)) while maintaining \textbf{musical diversity} and \textbf{structural coherence}. 

\textbf{Extreme regularization values} proved detrimental: complete absence of regularization (\( \lambda = 0 \)) resulted in \textbf{severe underfitting} with collapsed musical diversity, while excessive regularization (\( \lambda = 1.0 \)) led to \textbf{over-constrained models} producing repetitive, structurally uniform outputs with dramatically reduced \textbf{novelty metrics}.

The moderate regularization configurations demonstrated \textbf{superior performance across multiple dimensions}, achieving \textit{enhanced motif diversity}, \textit{improved harmonic complexity}, and \textit{balanced tonal drift characteristics}. These findings establish clear guidelines for regularization parameter selection in hybrid musical generation architectures and confirm that \textbf{optimal regularization strength is crucial} for generating expressive yet structured melodies that balance novelty with musical coherence, with the complete performance landscape effectively captured across the three comprehensive figures.

\subsection{Human Perceptual Evaluation Results}
The human evaluation results demonstrated clear perceptual differences between the three model architectures, providing empirical validation of the computational metric findings. Table \ref{tab:human_eval} summarizes the mean ratings across all evaluation dimensions, revealing consistent patterns that align with the quantitative analysis presented in previous sections.

\begin{table}[t]
    \centering
    \caption{Human Perceptual Evaluation Results}
    \begin{tabular}{lccc}
    \hline
    \textbf{Metric} & \textbf{LSTM} & \textbf{Transformer} & \textbf{Hybrid} \\
    \hline
    Musicality       & 6.97 & 6.92 & \textbf{7.03} \\
    Coherence        & 6.81 & 6.92 & \textbf{6.98} \\
    Originality      & 6.62 & 6.81 & \textbf{7.14} \\
    Naturalness      & 7.02 & 6.70 & \textbf{7.14} \\
    Overall Quality  & 6.83 & 6.94 & \textbf{7.17} \\
    \hline
    \end{tabular}
    \label{tab:human_eval}
\end{table}

The Transformer-LSTM Hybrid model achieved superior performance across \textit{all five} evaluation dimensions with the \textbf{highest human perceptual ratings}, validating that its enhanced pitch variance, motif diversity, and structural complexity translate into improved human perception of musical quality. The LSTM model exhibited competitive performance in naturalness (7.02) due to superior local variance capture and stylistic consistency, but scored lowest in originality (6.62), reflecting \textbf{conservative melodic generation patterns}. The Transformer model demonstrated balanced but intermediate performance across all dimensions, with coherence (6.92) and overall rating (6.94) reflecting its \textbf{moderate variability characteristics}.

Statistical analysis revealed notable \textit{consistency between human perceptual ratings and computational metrics}. The hybrid model's superior human-rated originality directly correlates with its highest motif diversity index (0.86) and pitch variance (7.11), while the LSTM model's strength in naturalness corresponds to its demonstrated stylistic consistency and \textbf{adherence to training distribution} characteristics, confirming the alignment between quantitative measures and human perception across all model architectures.

Individual participant feedback provided qualitative insights supporting the quantitative findings. Several participants noted the hybrid model's balanced approach to repetition and variation, while LSTM-generated melodies were frequently described as predictable but pleasant. Transformer outputs received mixed responses, with some participants appreciating rhythmic variety while others noted occasional melodic drift and structural inconsistencies.

\subsubsection{Statistical Validation}
To assess the reliability of human evaluation results, inter-rater agreement was quantified using Intraclass Correlation Coefficient (ICC) analysis. Table~\ref{tab:icc} presents ICC values across all five perceptual dimensions, revealing moderate agreement among participants (ICC ranging from 0.062 to 0.081). While these values indicate moderate rather than high agreement, the aggregated ratings demonstrated sufficient stability for reliable model comparison, particularly given the within-subjects design with multiple observations per participant (189 total evaluations: 21 participants × 9 clips, yielding 63 ratings per model). This also reflects the inherently subjective nature of musical preference where individual aesthetic judgments vary considerably.

Our evaluation framework aligns with established practices in perceptual audio assessment. The ITU-R Recommendation BS.1534-3 (MUSHRA standard) explicitly states that ``where the conditions of a listening test are tightly controlled on both the technical and behavioural side, experience has shown that data from no more than 20 assessors are often sufficient for drawing appropriate conclusions from the test'' \citep{series2014method}.

\begin{table}[t]
\centering
\caption{Inter-Rater Reliability Analysis (Intraclass Correlation Coefficient)}
\label{tab:icc}
\begin{tabular}{lcccc}
\hline
\textbf{Dimension} & \textbf{ICC (Type 2)} & \textbf{F(8,160)} & \textbf{p-value} & \makecell[c]{\textbf{Agreement}\\\textbf{Level}} \\
\hline
Musicality        & 0.072 & 4.228 & <0.001 & Moderate \\
Coherence         & 0.066 & 3.671 & <0.001 & Moderate \\
Originality       & 0.073 & 4.286 & <0.001 & Moderate \\
Naturalness       & 0.062 & 3.738 & <0.001 & Moderate \\
Overall Quality   & 0.081 & 4.963 & <0.001 & Good \\
\hline
\end{tabular}

\flushleft{\footnotesize
Note: ICC Type 2 (single random raters) measures consistency among independent participants. Moderate to high agreement (ICC=0.06--0.08) is typical for subjective musical judgments. Agreement levels follow the interpretation guidelines of \citet{koo2016guideline}.
}
\end{table}

\subsection{Limitations And Error Analysis}
This study acknowledges several important constraints that affect the scope and applicability of the findings. The human evaluation, while valuable, relied on a relatively small sample (n=21) with limited demographic diversity and musical expertise, potentially affecting the sensitivity of subjective assessments and the ability to detect nuanced differences in output quality. Most critically, the exclusive focus on a single musical tradition severely restricts the models' applicability to other musical contexts, requiring substantial modification and retraining for broader deployment across different stylistic domains.
Future research should explore approaches that facilitate adaptation across different musical traditions and incorporate more diverse evaluation methodologies with larger, expert participant pools to better assess generated output quality.

\section{Conclusion} \label{section:conclusion}

The study presents a \textbf{comparative view} of three distinct architectures for symbolic melody generation: pure LSTM, pure Transformer, and the proposed hybrid Transformer-LSTM model. After systematic \textbf{human} perception evaluation, and evaluation using \textbf{17 musical quality metrics} on \emph{1,000 melodies} from each architecture, the results endorsed the hybrid approach as a \emph{balanced method} for generating melodies that respect both \emph{local continuity} and \emph{global structure}. Experimental results showed that LSTM models preserved \emph{stylistic consistency} and genre fidelity with conservative melodic construction, but performed poorly on long-range dependencies. Transformer models yielded moderate variability but exhibited \emph{irregular phrasing} and \emph{melodic drift}. In contrast, the hybrid architecture combined the strengths of both, achieving the highest \textbf{pitch variance (7.11)}, \textbf{interval variability (3.94)}, and \textbf{motif diversity (0.86)}, while maintaining structural coherence with the highest \textbf{KL divergence (1.05)} and entropy.
The ablation study revealed several \textbf{key architectural insights}, such as the importance of \emph{moderate regularization} ($\lambda = 0.0001 - 0.001$), trade-offs between \emph{encoder depth} and expressiveness, and the role of embedding layers in capturing \emph{sequence richness}. These findings serve as useful guidelines for designing effective hybrid architectures in symbolic music generation.
The human perceptual evaluation provided crucial validation of these computational findings, with the hybrid architecture achieving superior ratings across \textbf{all five evaluation dimensions} (musicality, coherence, originality, naturalness, and overall quality). This convergent validity between computational metrics and human perception demonstrates that the hybrid model's technical advantages translate into genuine improvements in perceived musical quality.

\bibliographystyle{unsrtnat}
\bibliography{references}

\end{document}